# Performance Evaluation of Advanced Deep Learning Architectures for Offline Handwritten Character Recognition

Moazam Soomro
Department of Electronics and Power Engineering
Pakistan Navy Engineering College, NUST-PNEC
Karachi Pakistan,
moazam.soomro@pnec.nust.edu.pk

Muhammad Ali Farooq
Department of Electronics and Power Engineering
Pakistan Navy Engineering College, NUST-PNEC
Karachi, Pakistan
mali.farooq@pnec.nust.edu.pk

Rana Hammad Raza
Department of Electronics and Power Engineering
Pakistan Navy Engineering College, NUST-PNEC
Karachi, Pakistan
hammad@pnec.nust.edu.pk

*Abstract*— **This paper presents a hand-written character recognition comparison and performance evaluation for robust and precise classification of different hand-written characters. The system utilizes advanced multilayer deep neural network by collecting features from raw pixel values. The hidden layers stack deep hierarchies of non-linear features since learning complex features from conventional neural networks is very challenging. Two state of the art deep learning architectures were used which includes Caffe AlexNet [5] and GoogleNet models [6] in NVIDIA DIGITS [10]. The frameworks were trained and tested on two different datasets for incorporating diversity and complexity. One of them is the publicly available dataset i.e. Chars74K [4] comprising of 7705 characters and has upper and lowercase English alphabets, along with numerical digits. While the other dataset created locally consists of 4320 characters. The local dataset consists of 62 classes and was created by 40 subjects. It also consists upper and lowercase English alphabets, along with numerical digits. The overall dataset is divided in the ratio of 80% for training and 20% for testing phase. The time required for training phase is approximately 90 minutes. For validation part, the results obtained were compared with the ground-truth. The accuracy level achieved with AlexNet was 77.77% and 88.89% with Google Net. The higher accuracy level of GoogleNet is due to its unique combination of inception modules, each including pooling, convolutions at various scales and concatenation procedures.**

## I. Introduction

Deep learning is an emerging field that is progressing machine learning much closer to achieve higher classification accuracy levels. A successful deep learning model requires two essential aspects which include high computational power and a rich dataset. Deep neural networks (DNN) have been emphasized in pattern recognition and machine learning fields. As they are composed of many layers, DNNs can model much more complicated functions than simple neural networks.

Deep learning is providing desirable spectrum of great results across computer vision and pattern recognition problem domains. It plays a vital role in applications such as face recognition, image labelling, object detection, object classification and many more.

Handwriting recognition in pattern recognition is the capability of an algorithm to correctly predict the class label of the character in query. The input can be in two forms, through an image of the text which is known as offline approach and the other is writing on a tablet or a touch screen which is known as an online approach. In the online approach, much more information is available such as the pen trajectory and the image of the character itself, hence making recognition easier and faster.

Offline approach is more challenging as image of the handwritten character is obtained by either scanning or taking a picture of the document using a camera which subsequently results in noisy images, due to photometric, geometric and hardware constraints. This requires some pre-processing to be done on the images before feeding them to the algorithm. One of the preliminary and preprocessing task in character recognition to remove noise includes morphological operations such as thresholding and removal of textured background. These operations were carried out while creating our local dataset to remove noise.

There are numerous applications of identifying hand-written characters. It can be used to digitize old records in hospitals or offices. It can aid blind people by converting text to speech. Decoding handwritten scripts/notes while recycling PCBs. It can also be applied in post office for sorting letters area wise. Similarly, handwritten character recognition can also be used as an Optical Character Recognition (OCR) tool in non-standardized license plate recognition, which are still being observed in developing countries.

In this paper, we provide a comparative analysis of two state of the art deep learning algorithms for hand written character recognition using Caffe AlexNet [5] and GoogleNet [6]. Experimental results were compiled on two different datasets: Chars74K [4] dataset and another dataset was compiled locally.

The purpose of using two datasets is to make the training data richer and combined datasets can be used for more accurate testing of cases from another dataset also.

Brief background research is provided in next section that overviews the related work. Section III presents the implementation of the two state of the art deep learning frameworks for hand-written character recognition. Results are discussed in section IV and section V summarizes and gives concluding remarks.

## II. BACKGROUND RESEARCH

Handwritten character recognition has been an active area of research due to wide range of applications. Many different approaches and methods are used for precise and accurate classification of different numbers and alphabets. It includes Support Vector Machine (SVM), Naive Bayes Classifier and conventional Artificial Neural Networks (ANN).

Rahtu E. et al [2] implemented an affine invariant pattern recognition algorithm using multiscale auto convolution, in which they employed probabilistic interpretation of image functions. Their proposed work is for segmented objects and uses Fast Fourier Transform to reduce computational complexity. They approximated the affine transformations of the distortions present in the image, and suggested to be best suited where this approximation is possible. Results which were computed on recognition of binary images of English characters: A, B, C, D, E, F, G, H, and I, with added noise with some mis-classifications. The affine invariant moment with 2% binary noise and multiscale auto convolution with 6% binary noise resulting in high recognition errors. This approach depends on relatively complex mathematical computations, as compared to invariance learning [1].

Kamruzzaman and Aziz [3], in their research, offered a character recognition algorithm using neural network based double backpropagation method. The recognition is divided into two phases. Firstly, information as invariant features to rotation, translation, and scale are extracted out in the preprocessing phase. Later the neural network is trained on the computed features. They have achieved a classification rate of 97% on the testing images. The research cannot be confirmed as the neural network was never tested on a dataset of handwritten characters for robustness in real-world applications where slight anomalies can make the algorithm perform poor.

Deep learning denotes neural network architecture, but it has more than one hidden layer. These networks are inspired from biological structure which helps it overcome constraints and performance of single hidden layer networks. Deep neural network architecture benefits by a dispersed representation of features at each hidden layer, dissimilar features are extracted by neurons in each hidden layer, and multiple neurons are active simultaneously.

Deep learning architectures can be divided into three categories: Generative architectures, discriminative architectures and hybrid architectures. Generative architecture basically relies on unsupervised learning, where it performs clustering of the input data, examples of such networks are Deep Boltzmann Machines (DBMs), Deep Belief Networks (DBNs) etc. Discriminative architecture determines the class label of the input data, i.e. supervised learning, examples for this type of network include Deep Convolutional Networks, Deep Convex Network etc. Discriminative networks are used for fine tuning of generatively trained networks. Hybrid architectures are combination of generative and discriminative method, they are trained generatively and fine-tuned for deterministic motive [1].

Oyedotun et al [1], utilized Yoruba vowel characters for training and recognition. The dataset was divided in the ratio of 14,000 samples used to train the networks, 2,500 samples as the validation set, and 700 samples as test set for each invariance constraint. The networks that were trained includes the conventional Back Propagation Neural Network (BPNN), Denoising Auto Encoder (DAE), Stacked Denoising Auto Encoder (SDAE), and Deep Belief Network (DBN). The outcomes on the proposed dataset shows that DBN and SDAE have low error rates at relatively low noise levels, but their performances seem to degrade drastically from 7% and 10% noise densities respectively, while BPNN-1 was observed to have the best performance at 30% noise level.

## III. IMPLEMENTATION

In our research work we have utilized two datasets which are described in Table I

TABLE I. DATASETS

| Dataset | Attributes |
|---|---|
| Chars74K [4] | • The Chars74K dataset consists of 62 classes (0-9, A-Z, a-z), 7705 characters.<br>• It is obtained from natural images, 3410 hand drawn characters using a tablet PC, 62992 synthesized characters from computer fonts.<br>• This gives a total of over 74K images (which explains the name of the dataset). |
| Local Dataset | • The local dataset consists of 62 classes (0-9, A-Z, a-z), 4320 characters.<br>• The dataset was created by 40 subjects, each was handed out three sheets to write the alpha-numeric characters. Which were then scanned and cropped. |

For the proposed research work we have utilized two state of art deep neural networks which includes GoogleNet [6] and AlexNet [5]. The proposed model is based on supervised learning technique and uses deep Convolution Neural Networks (CNN). In CNN, a sample image is convolved with a filter kernel of (N x N) size to produce more refined output from raw pixels of image. The block diagram representation of the proposed research work is shown in Fig 1.

## A. Caffe AlexNet

In deep learning, we for the most part have utilized the Caffe AlexNet model for alpha numeric character recognition [5]. Alexnet model was trained using ImageNet [7] data. It contains over 15 million images from over 22000 different categories. The general structure of this system is fundamentally the same as CaffeNet model. They are a collective of 8 layers, the core 5 layers are convolutional layers, and the last 3 layers are fully-connected layers. The activation function utilized as a part of the hidden layer is "ReLU" layer, and the activation function of the output layer is SoftMax layer. Simply the system subtle elements are marginally unlike, that the grouping of the "norm" layer (lrn) and "pool" layer (max pooling) of the initial two layers is distinctive [8]. The general formula for Softmax layer is given in equation (1)

$$\sigma(z) = \frac{e^{z_j}}{\sum_{k=1}^{K} e^{z_k}} \; for \; j = 1, \ldots, k. \qquad (1)$$

where $\sigma(z)$ is the symbol for Softmax layer, $e^{z_j}$ is the exponential function and $\sum_{k=1}^{K} e^{z_k}$ is the summation function [9].

## B. GoogleNet

Another CNN model we utilized for alpha numeric character recognition is GoogleNet architecture [6]. One noteworthy trait of GoogleNet is that it is designed creatively, while the system consists of 22 layers deep network when checking just layers with parameters (or 27 layers if including the pooling layers). Another trait of GoogleNet is that another inception module was acquainted with CNN. The essential thought of inception module is to locate the ideal nearby structure and to reiterate it spatially. One of the principle and valuable parts of this architecture is that it expands the quantity of units at each stage expressively without an uncontrolled expansion in computational density.

For the proposed research work we have used both Caffe AlexNet and GoogleNet Networks in NVIDIA DIGITS [10] platform and compared their respective results which is shown in section (IV). The Fig 1 represents a flow diagram of hand written character recognition.

Fig 1 illustrates the stages of classification, when an input image of a character is given to the multi-layer deep neural network. Firstly, the image's tensor values are calculated which are extracted from the raw pixels of the image. Then these tensor values are fed to Caffe AlexNet and GoogleNet network. These networks perform different convolution operations at each layer of their architecture. Moreover, pooling layer and fully connected layers plays a vital role to produce the classification results. The function of pooling layer is to progressively reduce the spatial size of the representation which reduces the number of parameters and computation in the network, and hence to also control overfitting.

A fully connected layer takes all neurons in the previous layer (be it fully connected, pooling, or convolutional) and connects it to every single neuron it has. Fully connected layers are not spatially located anymore (can be visualized as one-dimensional), so there can be no convolutional layers after a fully connected layer. Finally, we evaluate a percentage score per class for which the system was trained. This percentage score denotes the confidence score. The class which comes up with the highest confidence score is considered to be the classifier's prediction.

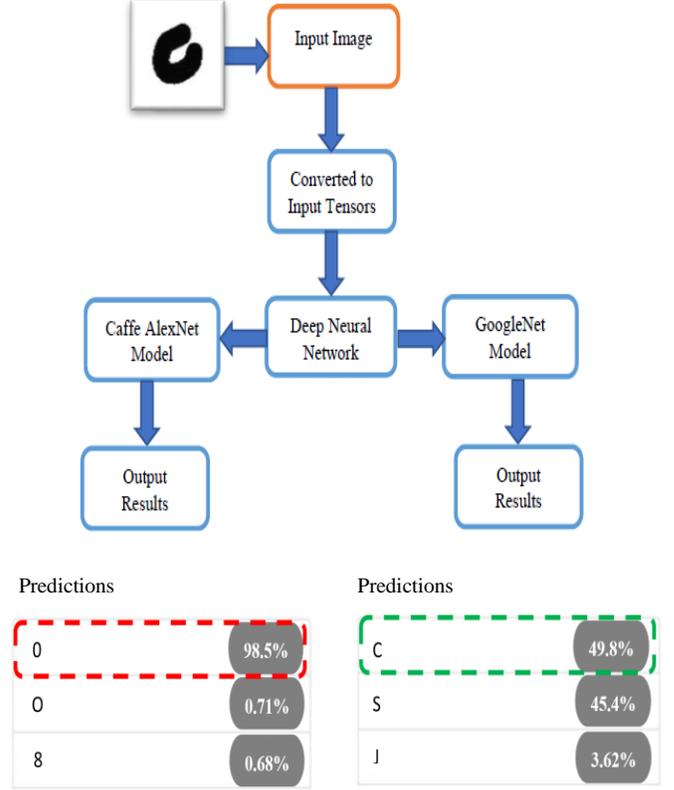

Fig. 1. Overall block diagram representation of dual deep neural networks

## IV. RESULTS

In this section, the proposed method is evaluated on NVIDIA DIGITS platform which comes with built-in models of Caffe AlexNet and GoogleNet networks for classification purpose. The NVIDIA Deep Learning GPU Training System (DIGITS) empowers deep learning under the control of specialists and researchers [10]. DIGITS can be utilized to quickly prepare the very precise Deep Neural Networks system (DNNs) for image classification, segmentation and object detection tasks.

The proposed system is implemented on HP Z440 machine having 32 GB of RAM and equipped with Nvidia Quadro K2200 Graphic Card. The NVIDIA Quadro K2200 transports excellent power-efficient 3D application performance. It has 4GB of GDDR5 GPU memory with fast bandwidth that allows us to create large, complex models, and a flexible single-slot form factor makes it compatible to fit in machine with limited space slot. It has clock size of 128 bits. The overall time required to train both the architectures was appx 90 minutes.

The Table II shows the sample images of the dataset created which have been used to train the classifier.

TABLE II. SAMPLE IMAGES OF CHARACTERS (A),(B), (C), (D), (E) & (F) REPRESENT CHARS74K, WHILE (G), (H),(I), (J), (K) & (L) ARE FROM LOCAL DATASET OF THREE DIFFERENT SUBJECTS.

| | | | | | |
|---|---|---|---|---|---|
| MS5J G2QP | ABPY 58HV | WIKO 27AE | US3N ZRTP | F641 DXQM | OLJ5 F9KW |
| (a) | (b) | (c) | (d) | (e) | (f) |
| ABCE 3ZXM | 7BOP QTXY | PGLO 83QI | AHNR 29DF | 7W09 IGRO | AZ53 2JFE |
| (g) | (h) | (i) | (j) | (k) | (l) |

To perform training and testing, the overall dataset is divided in the ratio of 80% for training and then rest of the data was used for testing. The training data was sub divided into three major parts which includes training, testing and validation with the respective ratios of 70%, 15% and 15%. When training the system, characters were preprocessed by using thresholding to remove the noise. Secondly, morphological operations such as Erode was performed on certain cases where there was discontinuity between the pixels, thus producing refined output. Both the networks were trained through 150 iterations with a learning rate of 0.001 as shown in Fig 2

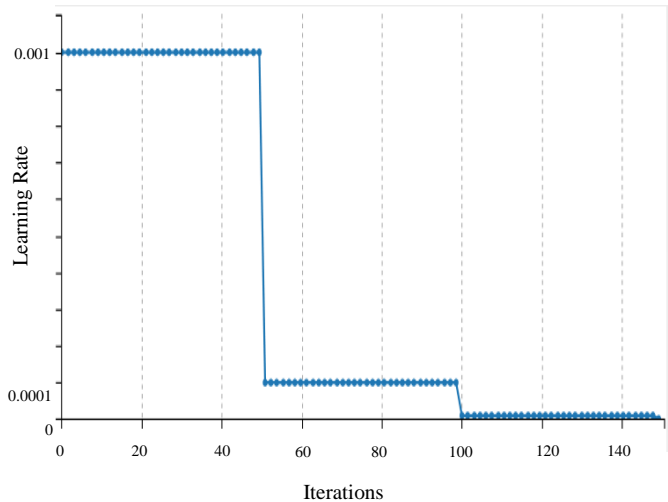

Fig. 2. Learning rate and number of iteration used for classification

The Fig 3(a) & 3(b) displays the overall training results obtained from Caffe AlexNet and Googlenet mutlilayer architecture.

### A. Results of Caffe AlexNet Network

This section will validate the results obtained on different test cases after the classifier is successfully trained. The classifier was tested on more than 500 characters amongst which few were selected for the research work.

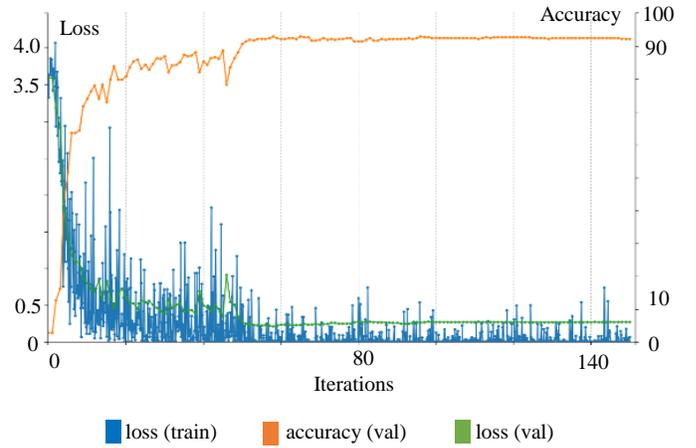

Fig. 3(a) Caffe AlexNet training graph with the validation accuray of appx 92.15% and validation loss of 0.27

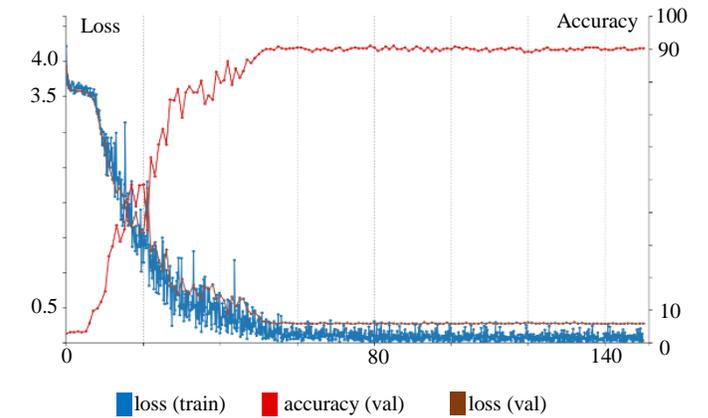

Fig. 3(b) GoogleNet training graph with the validation accuray of approximately 90.25% and validation loss of 0.28

### B. Results of GoogleNet Network

This section will validate the results obtained after the successful training of GoogleNet network.

### C. Comparasion of the results using Caffe AlexNet & GoogleNet model

After successful training and testing the certain cases using both the Caffe AlexNet and GoogleNet model the results were subsequently compared by calculating their respective accuracy levels using the equation (2).

$$Accuracy\ Level = \frac{No\ of\ Cases\ Correctly\ Classified}{Total\ Test\ Cases} \quad (2)$$

The Table III shows the visualization of statistical results obtained on each convolution layers using Caffe AlexNet network.

TABLE III. VISUALIZATIONS & STATISTICS OF CONV LAYERS FOR TEST CASE 'G', USING CAFFE ALEXNET NETWORK.

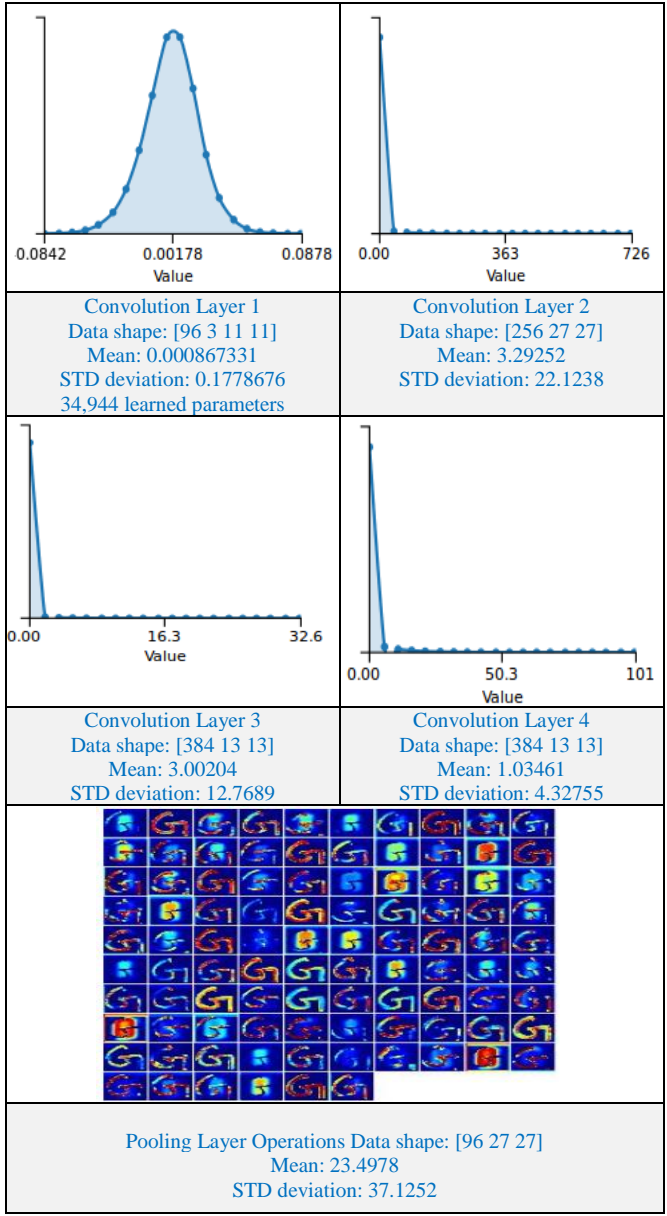

TABLE IV. RESULTS ACHIEVED ON THREE TEST CASES USING CAFFE ALEXNET & GOOGLENET ARCHITECTURES.

| Test Cases | Case 1 | | Case 2 | | Case 3 | |
|---|---|---|---|---|---|---|
| | G | | R | | 6 | |
| Caffe AlexNet Architecture | Predictions | | Predictions | | Predictions | |
| | G | 95.62% | R | 100% | 6 | 100% |
| | O | 1.57% | B | 0% | G | 0% |
| | S | 1.4% | 8 | 0% | H | 0% |
| GoogleNet Architecture | Predictions | | Predictions | | Predictions | |
| | G | 55.58% | R | 99.9% | 6 | 100% |
| | R | 11.87% | A | 0.01% | G | 0% |
| | D | 11.12% | 9 | 0% | H | 0% |

The Table IV shows the top three predictions of each character along with their confidence level scores using Caffe AlexNet and GoogleNet network.

For the proposed research work we have randomly selected eighteen different test cases. The overall results obtained on these eighteen cases using the two models are respectively shown in Table V. It should be noted that Table V shows the top prediction of test cases of each architecture along with its confidence score, while in Table IV we have shown the top three predictions for three cases from Table V.

The evaluated results shown in Table V reflects that the accuracy level using Caffe AlexNet is lower as we have got 14 true classification achieving the accuracy level of 77.77% whereas on the other hand using GoogleNet architecture we have achieved 16 correct classification results achieving the total accuracy level of 88.89%.

The higher accuracy level of GoogleNet shows that it has a significant diverse design when compared with AlexNet: it utilizes layers of inception modules, each including some pooling, convolutions at various scales and concatenation operations. It likewise utilizes 1x1 feature convolutions that work like feature selectors. Similarly, GoogleNet architecture provides 1x1 convolution block to reduce the number of features before the expensive parallel blocks also referred to as bottleneck layer. The inception module of GoogleNet [6] essentially performs collective convolution filter operations, which are arranged on same inputs. It also performs pooling procedures in parallel, and all the results are then concatenated. This enables the model to exploit multi-level feature extraction from each input thus having abundant feature values. Although using state of the art frameworks, we still encountered errors in similar characters like ('O' & '0'), ('7' & 'F'), ('5' & 'S'), and ('D' & 'O'). Such errors are still a challenging task in offline handwritten character recognition. A comprehensive comparison of the proposed work with other state of the art techniques has been shown in Table VI. The authors in [12] & [13] have utilized Chars74k dataset for handwritten digits and characters recognition using different algorithms, as shown in Table VI. According to authors in [13] the lower accuracy level of Alexnet is due to overfitting of model even with a dropout ratio of 0.8. Our proposed research work reflects the performance of GoogleNet with improved accuracy, when compared with [12] & [13].

V. CONCLUSION

In this paper, we provided a performance evaluation of two state of the art Deep Neural Networks (DNN) for the problem of hand written character recognition. For recognition purpose, we have utilized Caffe AlexNet [5] and GoogleNet [6] architectures in NVIDIA DIGITS for deep learning on the provided datasets. Training data was collected from two different datasets one of which was created locally for incorporating diversity, density and complexity. For preprocessing thresholding and morphological operations were applied on training local dataset to produce refined outputs.

After training, the system was successfully tested on characters (0-9, a-z, A-Z) to evaluate the system robustness. We showed that the GoogleNet architecture outperforms the Caffe AlexNet architecture on varying test cases, on the application of handwritten character recognition. The lower misclassification level of GoogleNet is due to its unique combination of inception modules, each including some pooling, convolutions at various scales and concatenation. Currently we have tested 18 different cases of upper & lower case letters along with numbers using varying complexity dataset, to evaluate the performance of both the architectures. For future work, we intend to assess more complex network, such as, ResNet [11] with substantial and comprehensive datasets.

TABLE V. OVERALL RESULTS ON EIGHTEEN DIFFERENT CASES USING CAFFE ALEXNET & GOOGLENET

| S. No | Test Cases | Confidence level (%) using AlexNet Architecture | Confidence level (%) using GoogleNet Architecture |
|---|---|---|---|
| 1 | G | 95.62% | 55.58% |
| 2 | H | 67.38% | 99.99% |
| 3 | I | 99.97% | 97.27% |
| 4 | J | 99.99% | misclassified with T |
| 5 | K | 100% | misclassified with H |
| 6 | L | 100% | 99.97% |
| 7 | R | 100% | 99.99% |
| 8 | T | 100% | 99.98% |
| 9 | U | 100% | 99.04% |
| 10 | S | 99.9% | 92.11% |
| 11 | v | 100% | 93.93% |
| 12 | X | 100% | 99.98% |
| 13 | 6 | 100% | 100% |
| 14 | 9 | misclassified with P | 75.18% |
| 15 | W | 100% | 100% |
| 16 | m | misclassified with W | 82.62% |
| 17 | Z | misclassified with 1 | 79.05% |
| 18 | P | misclassified with 9 | 80.33% |

TABLE VI. COMPARISON WITH OTHER STATE OF THE ART TECHNIQUES

| Method | Accuracy | Dataset | Classes |
|---|---|---|---|
| This Paper (GoogleNet) | 88.89% | Chars74K | 62 |
| This Paper (AlexNet) | 77.77% | Chars74K | 62 |
| Newell, Andrew J. et al. [12] | 80.00% | Chars74K | 62 |
| KNN [13] | 35.47% | Chars74K | 62 |
| Linear Classifier [13] | 30.15% | Chars74K | 62 |
| LeNet [13] | 45.36% | Chars74K | 62 |
| AlexNet [13] | 63.38% | Chars74K | 62 |
| Sundaresan, Vishnu et al. [13] | 71.69% | Chars74K | 62 |

ACKNOWLEDGMENT

We will like to thank Chars74K [4] research team for making the handwritten dataset available. We will also like to acknowledge NUST-PNEC students who helped us to create the local handwritten characters dataset.

REFERENCES

[1] Oyedotun, Oyebade K., Ebenezer O. Olaniyi, and Adnan Khashman. "Deep learning in character recognition considering pattern invariance constraints." International Journal of Intelligent Systems and Applications 7.7 (2015): 1J. Clerk Maxwell, A Treatise on Electricity and Magnetism, 3rd ed., vol. 2. Oxford: Clarendon, 1892, pp.68–73.

[2] Rahtu, Esa, Mikko Salo, and Janne Heikkila. "Affine invariant pattern recognition using multiscale autoconvolution." IEEE Transactions on pattern analysis and machine intelligence 27.6 (2005): 908-918.

[3] Kamruzzaman, Joarder, and Aziz d SM. "A neural network based character recognition system using double backpropagation." Malaysian Journal of Computer Science 11.1 (1998): 58-64

[4] de Campos, Teo, Bodla Rakesh Babu, and Manik Varma. "Character recognition in natural images." (2009).

[5] Krizhevsky, Alex, Ilya Sutskever, and Geoffrey E. Hinton. "Imagenet classification with deep convolutional neural networks." Advances in neural information processing systems. 2012.

[6] Szegedy, Christian, Wei Liu, Yangqing Jia, Pierre Sermanet, Scott Reed, Dragomir Anguelov, Dumitru Erhan, Vincent Vanhoucke, and Andrew Rabinovich. "Going deeper with convolutions." Proceedings of the IEEE conference on computer vision and pattern recognition. 2015.

[7] Deng, J., Dong, W., Socher, R., Li, L. J., Li, K., & Fei-Fei, L. "Imagenet: A large-scale hierarchical image database." Computer Vision and Pattern Recognition, 2009. CVPR 2009. IEEE Conference on. IEEE, 2009.

[8] Convulution Neural Networks
Web Link: http://cs231n.github.io/convolutional-networks/
(Last Accessed on 18 October, 2017)

[9] Bishop, Christopher M. Pattern recognition and machine learning. springer, 2006.

[10] NVIDIA DIGITS
Web Link: https://developer.nvidia.com/digits
(Last Accessed on 18 October, 2017)

[11] He, Kaiming, Xiangyu Zhang, Shaoqing Ren, and Jian Sun. "Deep residual learning for image recognition." Proceedings of the IEEE conference on computer vision and pattern recognition. 2016.

[12] Newell, Andrew J., and Lewis D. Griffin. "Multiscale histogram of oriented gradient descriptors for robust character recognition." Document Analysis and Recognition (ICDAR), 2011 International Conference on. IEEE, 2011.

[13] Sundaresan, Vishnu, and Jasper Lin. "Recognizing Handwritten Digits and Characters." , available on Stanford University Ftp server , 2015
Web Link: http://cs231n.stanford.edu/reports/2015/pdfs/
(Last Accessed on 18 October, 2017)